\documentclass[a4paper, 12pt]{article}

\usepackage{graphicx}
\usepackage{amsmath}
\usepackage{amssymb}
\usepackage[margin=20mm]{geometry}
\usepackage{natbib}
\usepackage{lineno}
\usepackage[hyphens]{url}

\begin{document}

\begin{center}
{\Large
Machine learning refinement of \textit{in situ} images acquired by low electron dose LC-TEM\\
}
\vspace{1em}
\textit{Running title}: Machine learning refinement on low dose LC-TEM
\vspace{1em}

Hiroyasu Katsuno$^{1,*}$, Yuki Kimura$^{2}$, Tomoya Yamazaki$^{2}$ and Ichigaku Takigawa$^{3,4}$ \\
\vspace{1em}
$^{1}$Emerging Media Initiative, Kanazawa University, Kakuma-machi, Kanazawa, 920-1192, Ishikawa, Japan\\
$^{2}$Institute of Low Temperature Science, Hokkaido University, Kita-19, Nishi-8, Kita-ku, Sapporo, 060-0819, Hokkaido, Japan\\
$^{3}$Institute for Liberal Arts and Sciences, Kyoto University, 302 Konoe-kae, 69 Konoe-cho, Sakyo-ku, Kyoto, 606-8315, Kyoto, Japan\\
$^{4}$Institute for Chemical Reaction Design and Discovery, Hokkaido University, N21 W10, Kita-ku, Sapporo, 001-0021, Hokkaido, Japan\\
$^{*}$Corresponding author: E-mail katsuno@staff.kanazawa-u.ac.jp, TEL +81-76-234-6935, FAX +81-76-234-6918
\end{center}

\vspace{1em}

\clearpage

\begin{abstract}
We study a machine learning (ML) technique for refining images acquired during \textit{in situ} observation using liquid-cell transmission electron microscopy (LC-TEM).
Our model is constructed using a U-Net architecture and a ResNet encoder.
For training our ML model, we prepared an original image dataset
 that contained pairs of images of samples acquired with and without a solution present.
The former images were used as noisy images and the latter images were used as corresponding ground truth images.
The number of pairs of image sets was $1,204$ and the image sets included images acquired at several different magnifications and electron doses.
The trained model converted a noisy image into a clear image.
The time necessary for the conversion was on the order of $10$\,ms, and
 we applied the model to \textit{in situ} observations using the software Gatan DigitalMicrograph (DM).
Even if a nanoparticle was not visible in a view window in the DM software because of the low electron dose,
 it was visible in a successive refined image generated by our ML model.
\end{abstract}

\vspace{1em}

\clearpage

\section*{Introduction}
\label{sec:intro}
Transmission electron microscopy (TEM) has been a powerful tool in the research field of materials science
 because TEM has revealed phenomena at the atomic scale
 such as an epitaxial Si(111) CoSi$_{2}$ interface \citep{haider1998electron},
 Au nanoparticles \citep{azubel2014electron},
 and order-disorder transition \citep{jeon2021reversible}.
Recent advances in the TEM technique have enabled the acquisition of images of a sample in a solution as well as under vacuum occurrence (liquid cell TEM, LC-TEM) \citep{de2011electron,de2019resolution}.
Because of interactions between electrons and a solution,
 some inevitable problems have arisen, such as blurred images \citep{holtz2013situ, nakajima2020time, yamazaki2021radiolysis}
and the occurrence of reactions due to the radiolysis of solutions \citep{schneider2014electron}.
Hence, LC-TEM observations and the corresponding analysis of the resultant data should be performed with these negative effects in mind.

One of the interesting nanoscale phenomena that occur in a solution is nonclassical nucleation.
The nucleation of crystals is the origin of all materials
 and the general understanding of this phenomenon is summarized in the classical nucleation theory \citep{Markov-book}.
The classical nucleation theory has only a single nucleation pathway in which molecules coalesce into crystals.
The justification of the classical nucleation theory has been confirmed by experiments on the epitaxial growth of metal or semiconductor materials \citep{Michely-book}. 
At the beginning of this century, new nucleation phenomena (called nonclassical nucleation phenomena) attracted many researchers,
 and various nucleation pathways have since been proposed \citep{deyoreo-Science-2015,Lee-PNAS-2016,Ishizuka-nanoscale-2017,vanDrisshe-Nature-2018}.
To clarify the nonclassical nucleation phenomena,
 it is necessary to directly visualize phenomena at the nanoscale; LC-TEM is a candidate tool for this purpose.
However, 
 TEM observation for an event in a solution is difficult
 because achieving 
 higher magnification, which results in a high electron dose, unless another observation condition is used, such as ptychography \citep{gao2017electron} and a higher accentuation voltage or an aberration corrector \citep{roserecognition}.
The radiolysis of a solution by the electron beam of a TEM can induce unexpected reactions.
In observations of samples that are sensitive to electron beams, only a low electron dose can be used,
 resulting in an unclear image.
Recent advances in machine learning methods have enabled the conversion of unclear microscopy images into clear images.
In the case of TEM, machine learning has also been successfully used for, for example, sparse coding \citep{olshausen1996emergence,leary2013compressed, stevens2014potential}
 and convolutional neural networks (CNNs) \citep{madsen2018deep,katsuno2022early,katsuno2022fast,kimura2022possible,krishna2023machine}.
The first CNN approach for pictures acquired by optical cameras mimics a sparse coding method, and its correspondence has been discussed \citep{dong2015image}.
A CNN can convert images in a short time, and real-time conversion has been realized \textit{in situ} observations by scanning TEM (STEM),
 where the STEM system was built to support remote autonomous experiments using edge computing and storage systems \citep{kalinin2023deep}.

In this paper,
 we investigate whether CNN can be used for image refinement during \textit{in situ} observations with standard equipment.
We prepared a handmade dataset for the training of a machine learning model.
The developed model was evaluated by qualitative evaluation for some model encoders,
 and the time necessary for image improvement was measured for certain image sizes.
We demonstrate visualization using a low electron dose for \textit{in situ} observation.
Using a standard function of the software Gatan DM,
 which is the industry standard software for image capture and analysis in (S)TEM,
 real-time conversion is achieved without the addition of any system.

\section*{Scheme of machine learning}

\subsection*{Collection of training data}
The dataset for training is the basis of machine learning and plays a crucial role and a good dataset will ensure that the model is properly trained \citep{kotsiantis2006machine}.
The model trains patterns from the dataset and uses these learned patterns to make predictions on new data.
The size and the diversity of the dataset are essential for managing overfitting and underfitting. 
One of the data collection methods for image improvement is to fabricate images using an image processing technique.
In training a model for super resolution,
 the dataset can be created by resizing the ground truth image.
Recently, the generation of simulated LC-TEM images was reported \citep{yao2020machine}.
The training dataset was fabricated by not only using a Gaussian filter to blur the image but also by using parameters on the experimental equipment.
Here, we adopted the method of making the dataset directly by using a transmission electron microscope because there are some unclear factors such as liquid thickness.
Our transmission electron microscope was equipped with a field-emission gun (JEM-2100F, JEOL, Tokyo)
 operated at an acceleration voltage of $200$\,kV, and a CMOS camera (One View IS, Gatan, Pleasanton, CA, USA).
The liquid cell consisted of a pair of silicon chips with an amorphous silicon nitride membrane of $50$\,nm thickness
as an observation window.
The LC-TEM holder (Poseidon, Protochips, Morrisville, NC, USA) is equipped with liquid injection ports,
 which were open in our operation.
A solution was injected into the liquid cell using a syringe when necessary.
Using our transmission electron microscope,
 we obtained a pair of images: one acquired in a solution as a noisy image and one acquired without a solution as a ground truth.
First, a sample was placed in the liquid cell (no injection of a solution)
 and images were acquired without a solution. 
The space between the amorphous membranes was likely filled with air.
A solution was then introduced into the liquid cell with a spacer thickness of $500$ nm, which controlled liquid thickness,
 and images were acquired in a solution.
The actual liquid thickness was much greater than the thickness of the spacer (see Supplementary Material).
This method is valid when the sample does not react with the solution.
We chose Au nanoparticles as a sample, and water as a solution.
When nanoparticles are encapsulated in a liquid cell,
 the particle positions change dramatically before and after the liquid injection.
That is, it is not possible to obtain corresponding image pairs without and with a solution.
This problem was avoided by placing the sample outside the silicon nitride membrane at the bottom side of the liquid cell.
It is known that images are different wether samples are placed on the top or bottom membrane \citep{zaluzec2015influence}.
In our case, images of nanoparticles placed on the top membrane were too blurry to be appropriate as a training dataset.
Nevertheless, automatic data collection was difficult because of the slight difference in the position of the nanoparticles.
The difference was probably a consequence of the bulging of the membrane with a solution being suppressed by the capillary force compared with the bulging of the membrane without a solution.
The pairs of acquired images were manually clipped to the region of interest,
 and the number of image pairs in the training dataset was $1,204$.  90\% of the pairs (1,083 pairs) were used for training and 10\% (121 pairs) were used for validation \citep{katsuno2023zenodo}.

\begin{figure}
\centering
\includegraphics[width=0.48\textwidth, bb=0 0 431 346, clip]{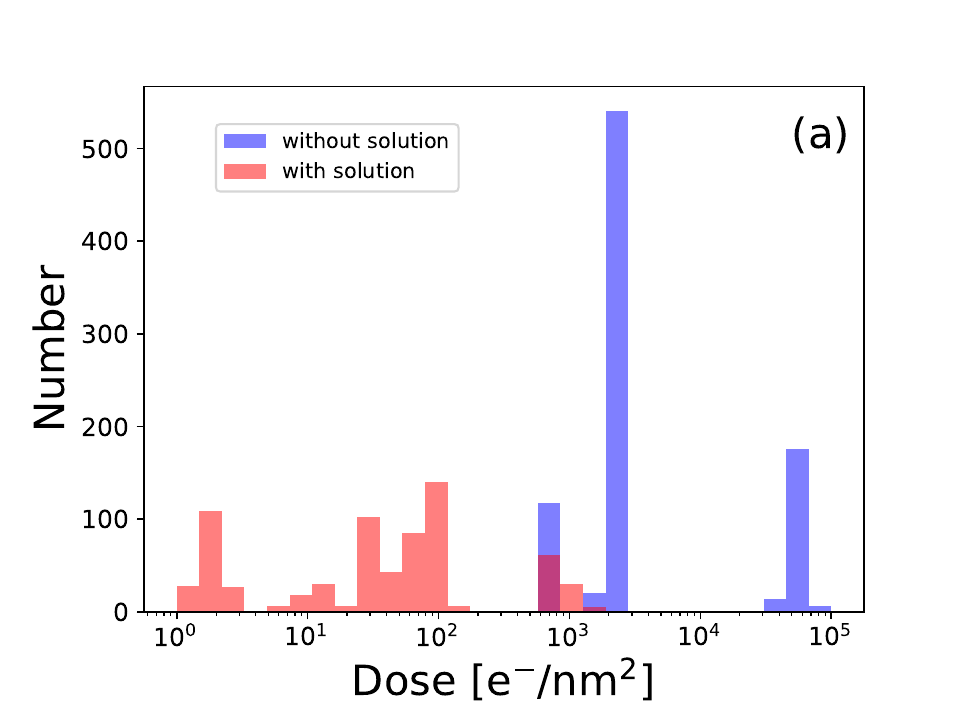}
\includegraphics[width=0.48\textwidth, bb=0 0 431 346, clip]{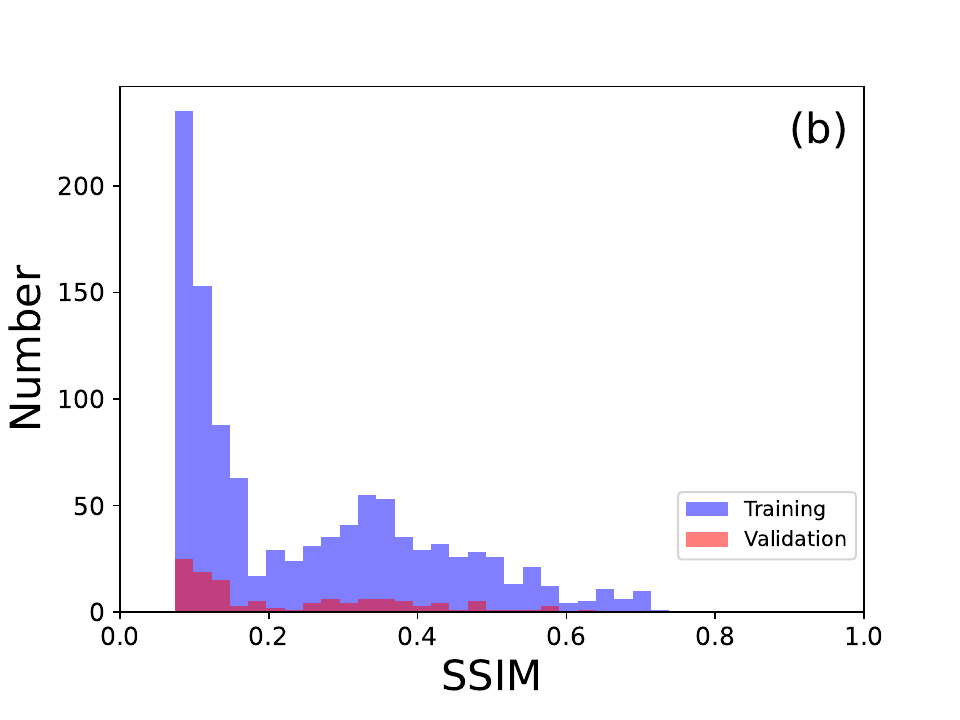}
\caption{
(a) Number distribution of the total dose of whole images acquired in the absence of a solution (blue) and in a solution (red).
(b) Number distribution of SSIM of pair images in training (blue) and validation (red)  dataset.
}
\label{fig:fig/14check_dose-dose/14check_dose-dose}
\end{figure}

The size of the acquired images was $4,096 \times 4,096$ pixels.
The typical magnifications used in the present study were $10,000\times$, $20,000\times$, and $100,000\times$.
The typical exposure time was $0.04$--$0.5$\,s for images acquired in a solution and $1$--$5$\,s for images acquired in the absence of a solution.
The lowest exposure time for $4,096 \times 4,096$ images in the software Gatan Digital Micrograph (DM) was $0.04$\,s,
 that is, 
 our dataset included the image that the user observed during \textit{in situ} observation and its total dose is approximately $1$--$1,000\,$e$^{-}$/nm$^{2}$ for solution samples.
The distribution of the total dose of each whole image in our dataset is shown in Figure~\ref{fig:fig/14check_dose-dose/14check_dose-dose}a.
The brightness differs by more than two orders of magnitude between pairs of images.
Figure~\ref{fig:fig/14check_dose-dose/14check_dose-dose}b shows the structural similarity index measure (SSIM) values [see equation~\ref{eqn:ssim}] of paired images in the training (blue) and validation (red) dataset.
About half of the paired images are between 0.1 and 0.2 and the other half of paired images are uniformly distributed between $0.2$ and $0.7$ in the training and validation datasets.

\subsection*{Machine learning model}
Our machine learning model was constructed with the U-Net architecture and the ResNet encoder using the segmentation package in Pytorch \citep{Iakubovskii:2019}.
U-Net is a simple network used in segmentation
 and includes a skip connection for maintaining the position of objects in the CNN \citep{ronneberger2015u,falk2019u}.
The ResNet encoder was developed for image recognition and has been demonstrated to effectively avoid the problem of vanishing gradients during training \citep{he2016deep}.
After the final convolution layer of the U-Net architecture, a sigmoid function was applied so that the intensity value of each pixel was in the range $0$--$1$
 because the output images were quantitatively compared to images with their data range of $0$--$1$.

\begin{figure*}
\centering
\includegraphics[width=\textwidth, bb=100 50 860 540, clip]{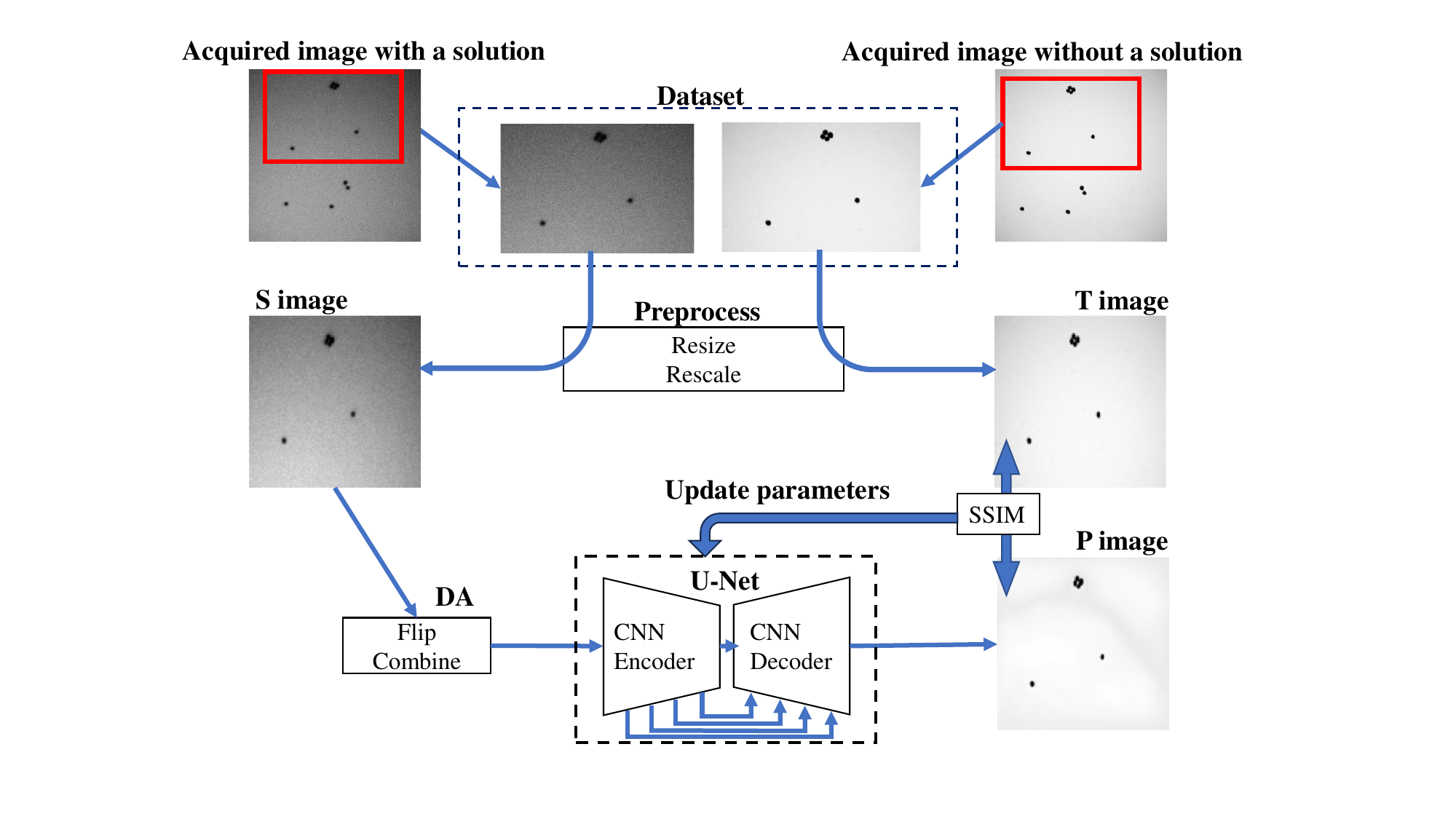}
\caption{
Schematics of making dataset and training of our machine learning model for predicted images (P image) from noisy images (S image) acquired in a solution.
The model parameters are updated
 by using the loss function SSIM shown in equation~(\ref{eqn:ssim})
 so that the P image is identical to the corresponding T image.
}
\label{fig:model}
\end{figure*}

\subsection*{Training}
A schematic of the training procedure is shown in Figure~\ref{fig:model}.
Before 
 a solution-acquired image of the training dataset was input into the machine learning model,
 the image was resized to $512 \times 512$ 
by using the resize function with the option cv2.INTER\_AREA provided by the OpenCV package \citep{opencv_library}
and its intensity data were rescaled to $0$--$1$.
By comparing the case of $512\times 512$ images of the camera output and the case of converting $4,096\times 4,096$ images of camera output to $512\times 512$ images,
 we confirmed that no artifacts are generated by the preprocess.
Hereafter, we refer to a preprocessed image as an S image.
The S images were then randomly flipped for data augmentations (DAs).
In addition, 
 a single image was sometimes created from four images resized to $256 \times 256$.
The model parameters were obtained using the Adam optimizer \citep{kingma2014adam} with a learning rate of less than $10^{-4}$.
The loss function was constructed using the SSIM \citep{wang2004image}
 with a predicted image (called P image) and corresponding ground truth image (called T image).
The SSIM is one of the methods for evaluating the perceived quality of digital images.
The SSIM can be calculated from two images $x$ and $y$ of a common region,
\begin{align}
\textrm{SSIM}(x,y)=\frac{ (2\mu_{x}\mu_{y}+c_{1}) (2\sigma_{xy}+c_{2}) } { (\mu_{x}^{2}+\mu_{y}^{2} + c_{1}) (\sigma_{x}^{2}+\sigma_{y}^{2}+c_{2})},
\label{eqn:ssim}
\end{align}
where $\mu_{x}$ and $\mu_{y}$ are the averages of $x$ and $y$, respectively.
$\sigma_{x}^{2}$ and $\sigma_{y}^{2}$ represent the variances of $x$ and $y$, respectively
 and $\sigma_{xy}$ represents the covariance of $x$ and $y$.
Parameters $c_{1}$ and $c_{2}$ are introduced to stabilize the division
and are set to $0.01$ and $0.03$, respectively.
Both values are commonly used such as default values in the scikit-learn package \citep{scikit-learn}.
The SSIM of a pair of images is obtained over the entire image with the common region shifted
 and indicates the average of equation~\ref{eqn:ssim}.
If two images are identical, the SSIM becomes $1$, which is the maximum value.
If there is no similarity between the two images, the SSIM value is $0$.
Using the SSIM, the loss function is defined as 
\begin{align}
\textrm{Loss function} = 1 - \frac{1}{N}\sum_{\textrm{region}} \textrm{SSIM}(x,y),
\label{eqn:lossfunc}
\end{align}
where $N$ represents the number of common regions whose size is $11\times 11$,
which is the standard value in the scikit-image package \citep{scikit-learn}.
We also used $3\times 3$ and $39\times 39$ as the common size, which resulted in no significant change.

All training was conducted on a Linux machine with a 10-core Intel i9-9900X 3.5GHz CPU and an NVIDIA Quadro RTX $8000$ graphics card.
The training was almost converged at $1,000$ epochs, but the training was performed up to $2,000$ epochs.
The overfitting is a topic of ongoing discussion in the field of machine learning \citep{belkin2018overfitting,bartlett2020benign,advani2020high}.
To suppress the overfitting, during the training, the data augmentation with the randomness was performed and we confirmed that the value of the loss function for validation did not increase.
The training took about $50$ minutes when using the ResNet-$18$ encoder without a pre-trained model such as the ImageNet data.
The code was available on the website \citep{katsuno2023github}.

\section*{Results and discussion}

\subsection*{Examples of refined images}

\begin{figure*}
\centering
\includegraphics[width=\textwidth, bb=50 0 910 540, clip]{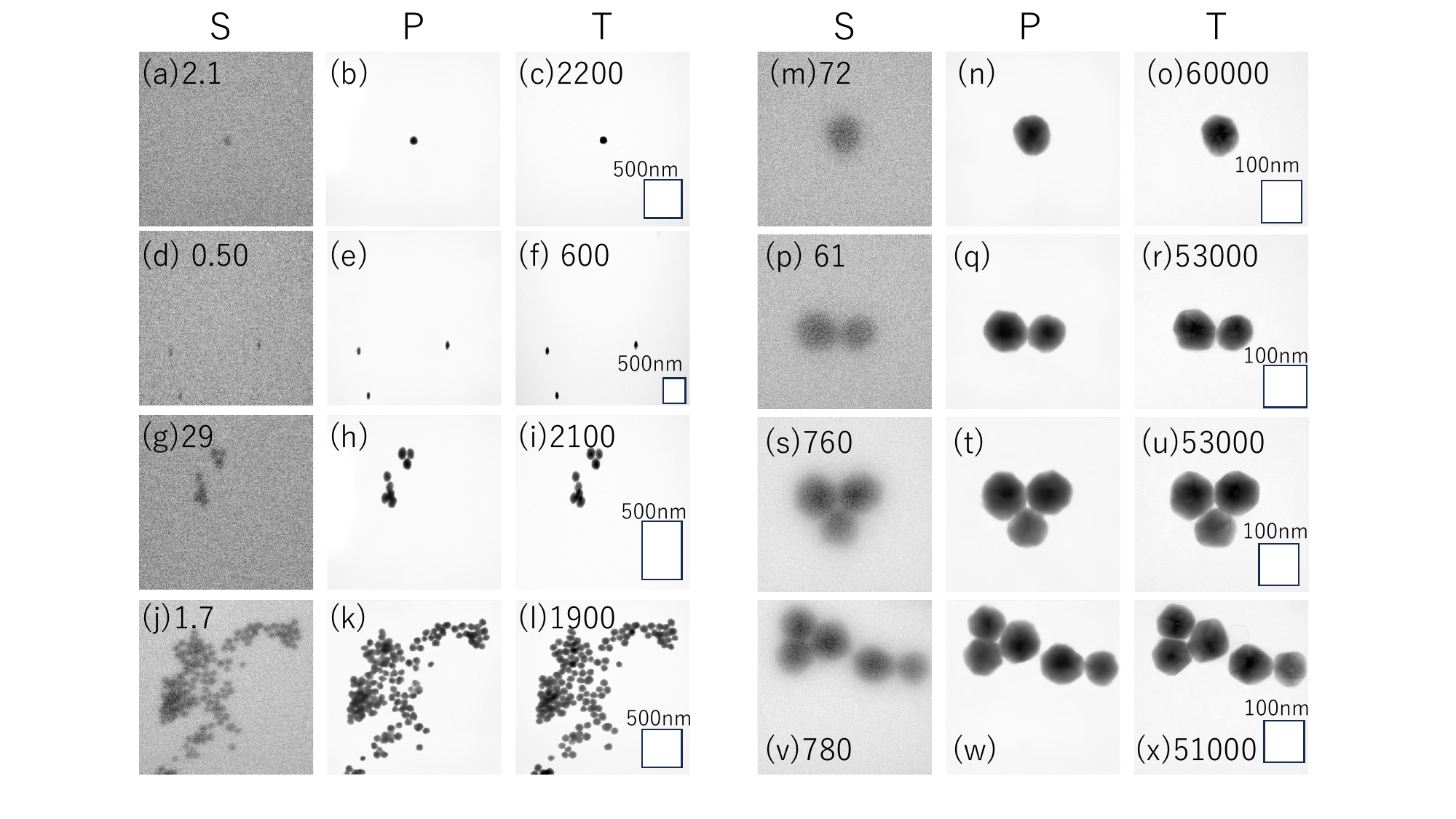}
\caption{
 Examples of S, P, and T images.
These images are not used for training.
S and T images were preprocessed images of the original images.
All objects are Au nanoparticles with a diameter of $100$\,nm.
White boxes in each T image represent a $500$ nm scale for (c), (f), (i) and (l), and 
 a $100$\,nm scale for (o), (r), (u) and (x).
The magnitude of the scale of the width differs from that of the height because clipped images are not square.
The number on the right side of the label on the S and T images represents the total dose expressed in e$^{-}$/nm$^{2}$.
}
\label{fig:fig/figs_p8}
\end{figure*}

Figure~\ref{fig:fig/figs_p8} shows examples of
 preprocessed images of a sample in a solution (S images),
 output images improved with our machine learning model (P images),
 and images of a sample in the absence of a solution (T images).
These images were not used for training.
All of the samples in the images are Au nanoparticles with a diameter of $100$\,nm.
The number shown on the right side of the label in each figure is the total dose.
Even with an electron dose of $1/1000$, we obtained P images comparable to T images.
In the low-magnification images in Figures~\ref{fig:fig/figs_p8}a and \ref{fig:fig/figs_p8}d,
 a nanoparticle becomes relatively small in the image; distinguishing between nanoparticles and noise in this case is difficult.
Our model successfully reproduces the positions of the particles without missing them (Figures~\ref{fig:fig/figs_p8}b and \ref{fig:fig/figs_p8}e).
The refined images (Figures~\ref{fig:fig/figs_p8}b and \ref{fig:fig/figs_p8}e) are comparable to the corresponding T images (Figures~\ref{fig:fig/figs_p8}c and \ref{fig:fig/figs_p8}f)
 and do not produce extra particles.
The P images (Figures~\ref{fig:fig/figs_p8}h and \ref{fig:fig/figs_p8}k)
 corresponding to the images in Figures~\ref{fig:fig/figs_p8}g and \ref{fig:fig/figs_p8}h,
 in which many particles are observed but the boundaries of particles are blurred and unclear,
are equivalent to the T images in Figures~\ref{fig:fig/figs_p8}h and \ref{fig:fig/figs_p8}k, respectively.
Image acquisition with high magnification brings additional benefits.
In the S images,
 the particles appear spherical because of the blurring caused by the solution (Figures~\ref{fig:fig/figs_p8}m, \ref{fig:fig/figs_p8}p, \ref{fig:fig/figs_p8}s and \ref{fig:fig/figs_p8}v).
In the P images,
 objects with polygonal shapes are observed (Figures~\ref{fig:fig/figs_p8}n, \ref{fig:fig/figs_p8}q, \ref{fig:fig/figs_p8}t and \ref{fig:fig/figs_p8}w),
 in agreement with the images acquired in the absence of a solution (Figures~\ref{fig:fig/figs_p8}o, \ref{fig:fig/figs_p8}r, \ref{fig:fig/figs_p8}u and \ref{fig:fig/figs_p8}x, respectively).
We found that the shape restoration is effective even when particles are located adjacent to each other.

\subsection*{Quality of improvement}
\begin{table}
\centering
\caption{
PSNR and SSIM of pairs of a P image and T image for the combination of loss functions and ResNet encoders.
`Original' represents the pairs of an S image and a corresponding T image.
}
\vspace{1em}
\begin{tabular}{cccc}
\hline
Loss function & Encoder  & PSNR & SSIM \\
\hline
\hline
SSIM           & ResNet-$18$ & $29.52$ & $0.85$ \\
SSIM           & ResNet-$34$ & $29.45$ & $0.85$ \\
SSIM           & ResNet-$50$ & $29.27$ & $0.85$ \\
SSIM           & ResNet-$101$& $30.12$ & $0.85$ \\
SSIM           & ResNet-$152$& $29.87$ & $0.85$ \\
$L_{1}$        & ResNet-$18$ & $29.99$ & $0.85$ \\
$L_{2}$        & ResNet-$18$ & $29.65$ & $0.85$ \\
\hline
\multicolumn{2}{c}{(Original)} & $8.48$ & $0.08$ \\
\end{tabular}
\label{tab:PSNRSSIMforvalid}
\end{table}

Table~\ref{tab:PSNRSSIMforvalid} shows the quantitative evaluations of our model
 in terms of the peak signal-to-noise ratio (PSNR) and the SSIM.
Both parameters are one of the most frequently used evaluations in machine learning models such as \citep{chen2018learning}.
The PSNR is a logarithm of the inverse of the mean square error at the pixel level.
A high PSNR value indicates that the two images are similar to each other. 
The SSIM is the same as defined in equation~\ref{eqn:ssim}, 
 and the model parameters are trained such that the SSIM is equal to $1$.
PSNR and SSIM are widely used as image quality degradation evaluation methods.

The S images are very noisy,
 and finding Au nanoparticles at low magnification is difficult.
After our model is applied, S images are converted to P images.
When the encoder ResNet-$18$ is used,
 the average PSNR and SSIM for P images and the corresponding T images are $29.52$ and $0.85$, respectively.
Because noise was reduced, the P images are very similar to the corresponding T images.
We have evaluated other ResNet encoders listed in Table~\ref{tab:PSNRSSIMforvalid}.
Although large encoders are expected to obtain better results,
 we observed no significant change compared with the results obtained using the smallest encoder, ResNet-$18$.
Our results show that the smallest encoder is sufficient in our case of improving grayscale images with $16$-bit expression in each pixel.

We also evaluated different types of the loss function.
The $L_{1}$ ($L_{2}$) loss function is the mean absolute (squared) error and the absolute (squared) difference in the intensity at each pixel.
The obtained model parameters were comparable to those in the case  where the SSIM was used.
As already pointed out in previous studies \citep{chen2018learning},
 changing the loss function makes no significant difference in this framework
 although it affects the magnitude of the gradient during training.

The 'Original' column represents the values of the pairs of S and corresponding T images.
The average PSNR and SSIM in the dataset of S and T images are obtained as $8.48$ and $0.08$, respectively.
Our model is able to improve the image beyond simple data rescaling performed in the preprocessing.

\subsection*{Time necessary}

\begin{table}
\centering
\caption{
Conversion time of one image for various sizes of an image.
The time for a GPU includes the data transfer between the CPU and GPU.
}
\vspace{1em}
\begin{tabular}{lccc}
\hline
Processor & $256\times 256$ & $512\times 512$ & $1,024\times 1,024$ \\
\hline
\hline
Intel i7-9700           &  $106$\,ms &  $240$\,ms & $787$\,ms\\
NVIDIA Geforce GTX 1650 & $12.1$\,ms & $31.4$\,ms & $133$\,ms\\
NVIDIA RTX A2000        &  $8.0$\,ms & $13.3$\,ms & $40.8$\,ms\\
NVIDIA Quadro RTX 8000  &  $4.3$\,ms &  $5.9$\,ms & $19.1$\,ms\\
\hline
\end{tabular}
\label{tab:conversiontime}
\end{table}

We investigated the conversion time required for the conversion of an S image by our machine learning model.
The model encoder for this test was ResNet-$18$.
For the comparison, one CPU (Intel i7-9700) and three different GPUs (NVIDIA Geforce GTX 1650, NVIDIA A2000, and NVIDIA Quadro RTX 8000) were used.
The results are summarized in Table~\ref{tab:conversiontime}.
When the image size is smaller, the conversion time becomes shorter.
When an Intel i7-9700 was used,
 the conversion time was greater than $100$\,ms even for the $256 \times 256$ images.
Although there is no problem for normal use,
 this long conversion time reduces the temporal resolution of \textit{in situ} observation using DM.
When GPUs were used,
 the conversion time for the $1,024\times 1,024$ image was four times longer than that for the $512\times 512$ image.
This result is reasonable because the image data is four times larger.
When the most powerful processor was used,
 the time to process a $256 \times 256$ image was reduced to only $70\%$ of the time to process a $512 \times 512$ image.
By contrast,
 for the least powerful processors, the processing time for a $256 \times  256$ image was less than $50$\% of the time to process a $512 \times 512$ image.
That is, the more powerful processors tended to offer less benefit from a reduction of the image size.
We chose an image size of $512 \times 512$ for the conversion under \textit{in situ} observation.

The time required depends largely on machine power.
Therefore, the image might be improved within a short time if a machine dedicated to image improvement is used.
Even in the current state, we found that the refinement of images of \textit{in situ} observation is possible
 by simply integrating a reasonably priced GPU into a PC used to control a transmission electron microscope, without building a dedicated machine for image refinement.

\subsection*{Test for \textit{in situ} observation}
\begin{figure*}
\centering
\includegraphics[width=0.45\textwidth, bb=0 0 461 346,clip]{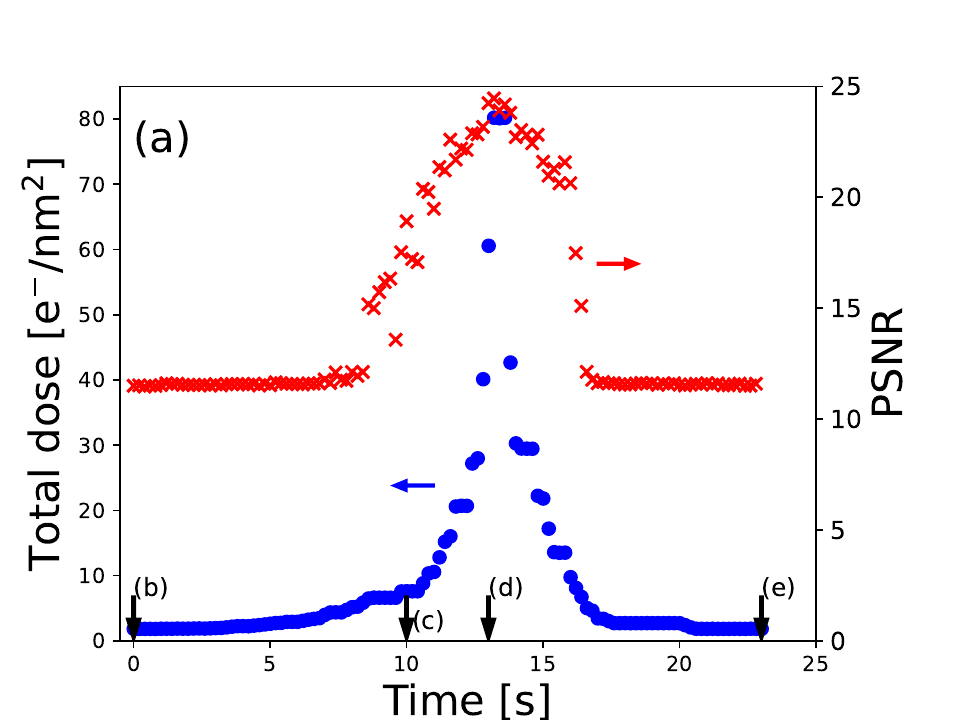}
\includegraphics[width=\textwidth, bb=0 0 960 540,clip]{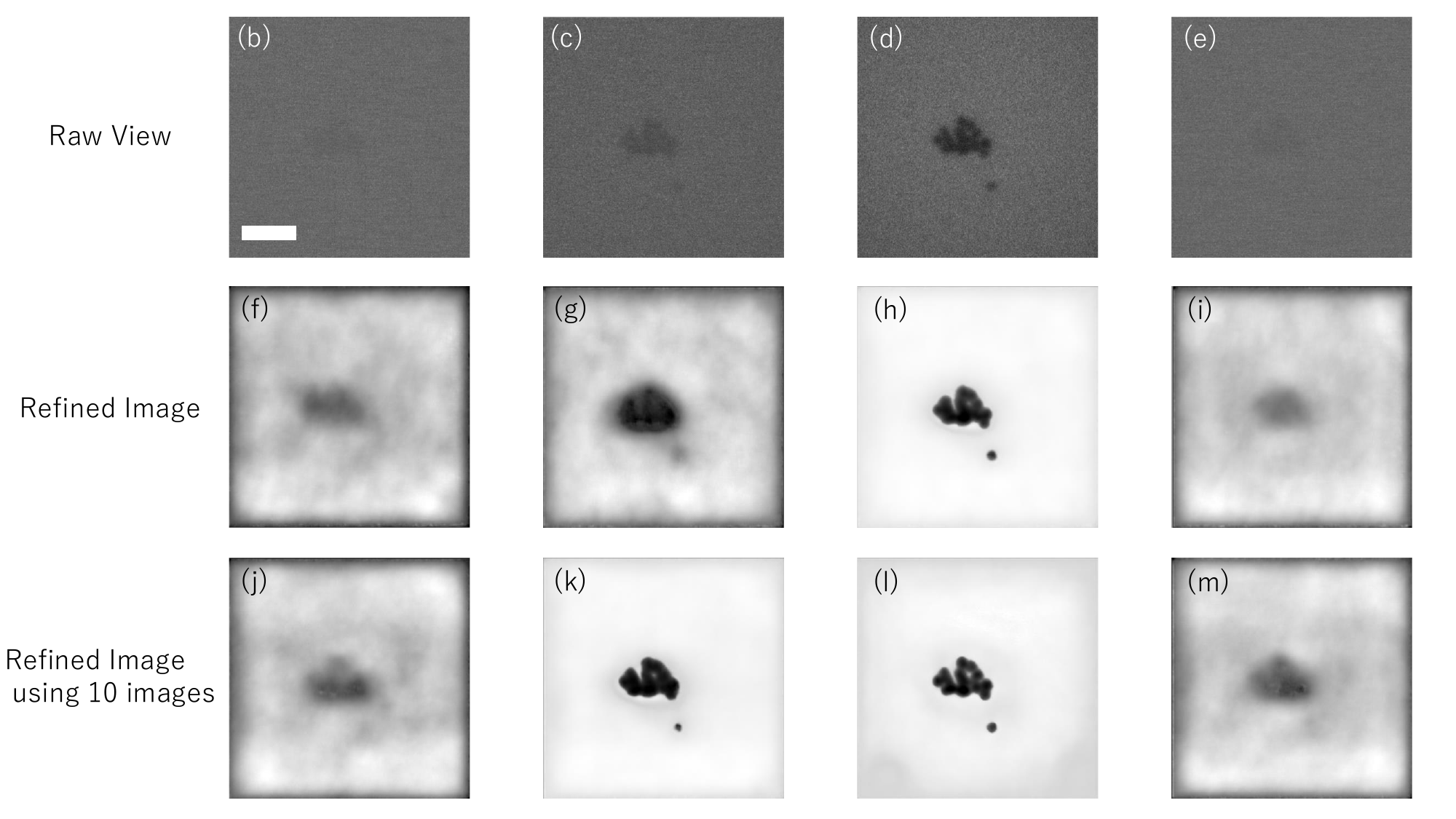}
\caption{
 Examples of images for P images under \textit{in situ} observation using the software Gatan DM.
(a): Time change of 
the total dose (filled circles)
 and 
PSNR (crosses) with respect to (l).
 for each image.
(b)--(e): Images acquired with an exposure time of $0.01$\,s at $t=0, 10, 13$
 and
 $23$\,s, as shown by
 down
 arrows in (a).
(f)--(i): Images refined from (b)--(e), respectively, by our machine learning model.
(j)--(m): Each of these images was created by combining the images (b)--(e) and nine images preceding them, and then refining them by our machine learning model.
The scale bar shown in (b) represents $500$\,nm.
}
\label{fig:doserate}
\end{figure*}

To test whether our machine learning model is effective for \textit{in situ} observations,
 we observed Au nanoparticles in a solution.
When an image with the current maximum size of $4,096\times 4,096$ was acquired, the system became slow when outputting acquired images to the view window.
When a $2,048\times 2,048$ image was acquired with an exposure time of $0.01$\,s,
 the appearance in the view window showed the same temporal resolution, as if the machine learning model did not work.
When a $2,048\times 2,048$ image with a magnification of $20,000\times$ was acquired,
 the images were converted to $512\times 512$ and the data were rescaled,
 corresponding to the preprocessing shown in Figure~\ref{fig:model}.
Preprocessed images were input to the machine learning model, and the output images were displayed.
All processes were performed on the software Gatan DM and NVIDIA RTX A2000 GPU.
Although the time necessary for the conversion is $13.3$\,ms from Table~\ref{tab:conversiontime}, 
the time necessary from a camera output to display a refined image is $63$\,ms.
Therefore, the time necessary for drawing in Gatan DM with python script is about $50$\,ms.
Although the recording was made with $100$ fps, the images are displayed about $10$ Hz, we had no problems with the slow display in actual \textit{in situ} TEM observation.

Because our goal is to recognize the presence or absence of samples in a solution at very low electron doses,
 we used the Gatan DM software to compare the raw view images and refined images collected at various electron doses.
The change in the electron dose is shown in Figure~\ref{fig:doserate}a, where the magnitude of the electron dose was measured in the background.
The left $y$-axis shows the dose rate per image.
The right $y$-axis shows the PSNR of each image when Figure~\ref{fig:doserate}l is used instead of the S image.
Some of the raw images viewed in Gatan DM are shown in Figures~\ref{fig:doserate}b--\ref{fig:doserate}e.
The time corresponding to each image is indicated by the arrow in Figure~\ref{fig:doserate}a.
The electron dose represented in Figure~\ref{fig:doserate}b is approximately 1\,e$^{-}$/nm$^{2}$, where the dose rate corresponds to 100\,e$^{-}$/nm$^{2}$ s.
We could not find Au nanoparticles in Figure~\ref{fig:doserate}b.
In the refined image in Figure~\ref{fig:doserate}f, an object is observed at the center.
The electron dose represented in Figure~\ref{fig:doserate}c is about 10\,e$^{-}$/nm$^{2}$, and an object is visible.
In the refined images in Figure~\ref{fig:doserate}g, an object is clearly observed
 but the shape restoration of the samples could not be realized.
The electron dose represented in Figure~\ref{fig:doserate}d is approximately 80\,e$^{-}$/nm$^{2}$, which is in the range of the electron dose used in our training dataset (see Figure~\ref{fig:fig/14check_dose-dose/14check_dose-dose}a);
 some particles of well-defined size are visible although the image is very noisy.
The refined image in Figure~\ref{fig:doserate}h provides the same information for samples but without noise.
After acquiring the image in Figure~\ref{fig:doserate}d,
 we again acquired the image in Figure~\ref{fig:doserate}e 
 whose refined image in Figure~\ref{fig:doserate}i shows the object at the center as dose Figure~\ref{fig:doserate}f.
Because the electron dose rate
 we usually use during the search for nanoparticles in samples is approximately $10,000$\,e$^{-}$/nm$^{2}$\,s,
 we can detect an object with an electron dose of $1/100$.

Our model was trained well without over-fitting
 because the model can refine the images in Figures~\ref{fig:doserate}c and d.
This behavior was expected, as shown in Figure~\ref{fig:fig/figs_p8}.
In these dark images, we observe nanoparticles.
Using this model,
 we succeeded not only in brightening but also in visualizing images that are difficult to judge with the human eye (Figures~\ref{fig:doserate}b and \ref{fig:doserate}e).
In an actual adjustment,
 the presence of an object can be seen more clearly if the previous nine images are integrated and refined (Figures~\ref{fig:doserate}j--\ref{fig:doserate}m) rather than simply refined one by one.
The total electron dose in the input images corresponding to Figures~\ref{fig:doserate}j--\ref{fig:doserate}m is approximately ten times greater than those in Figures~\ref{fig:doserate}b--\ref{fig:doserate}e.
More discriminative images can be obtained by adding the preceding images that were already acquired for the image.

We here consider the other aspect of the effect of electron-beam irradiation of a sample.
After the TEM observation begins,
 water in the sample will reach a steady state within 10$^{-3}$\,s under the TEM observation conditions.
The action of ionization is inevitable.
In particular, it affects the pH of the water \citep{schneider2014electron}.
The dose rate during an ordinary TEM observation is approximately 10$^{7}$ Gy/s,
 and the pH does not change when the initial pH of a liquid is less than 4.
Because our work has reduced the total dose required for TEM visualization by two orders of magnitude,
 we can visualize phenomena in a solution with a pH less than 6 without substantially changing its pH.

Currently,
 our method can reveal where objects are located but cannot reproduce their structure in detail.
Our model tends to binarize the image.
We were not sure why output images would be binarized.
One possibility is a sigmoid function put just before the output to avoid negative values.
However, the sigmoid function is probably not the direct cause,
 since another study \citep{katsuno2022fast} has shown that even cases without a sigmoid function also tend to be binarized.
Another possibility is the small size of the dataset.
Although we assume that a larger dataset is necessary to reproduce a detailed structure,
 large datasets are difficult to create in a straightforward manner. 
It would be necessary to create a pseudo-dataset in a way that uses empirical functions such as the work \citep{yao2020machine}.
If large datasets could be created,
 other models such as generative adversarial networks (GANs) \citep{goodfellow2014generative,goodfellow2020generative} would be better suited for use with our method.
We attempted to use GAN, but the result was unstable under the current situation with a small dataset;
 we therefore did not adopt it in the present work.

The Au nanoparticles in this manuscript are located on the outside of the lower membrane of the cell.
We have tested this model on several images in which the particles are placed inside the cell,
 with moderately good results.
This result means that our proposed machine learning model can be used for ordinary LC-TEM observation.

\subsection*{Summary}
We developed a method to improve \textit{in situ} TEM images using machine learning techniques.
Our model was constructed with the U-Net architecture and the ResNet-$18$ encoder.
The training dataset was prepared from experimental sets of noisy images acquired in a solution and corresponding truth images acquired in the absence of a solution.
The PSNR and the SSIM of the improved images were $29.52$ and $0.85$, respectively.
Our model was applied using the software Gatan DM, which controls the CMOS camera of a transmission electron microscope.
The refined images enabled us to observe nanoparticles that could not be observed in the view window without almost loss of temporal resolution.
Although our dataset was the image of particles outside of the cell, 
 machine learning model trained using this dataset would be applicable to ordinary LC-TEM observation in which samples were located inside the cell.

\section*{Acknowledgments}
This work was supported by JSPS KAKENHI Grant Numbers 20H05657 and 21K03379,
 the Grant for Joint Research Program of the Institute of Low Temperature Science, Hokkaido University (23G016),
 and the Grant for the Public Foundation of Chubu Science and Technology Center.

\bibliographystyle{abbrvnat}

\end{document}


\begin{center}
{\Large
Supplementary Material for
``Machine learning refinement of \textit{in situ} images acquired by low electron dose LC-TEM"\\
}

\vspace{1em}

Hiroyasu Katsuno$^{1,*}$, Yuki Kimura$^{2}$, Tomoya Yamazaki$^{2}$ and Ichigaku Takigawa$^{3,4}$ \\
\vspace{1em}
$^{1}$Emerging Media Initiative, Kanazawa University, Kakuma-machi, Kanazawa, 920-1192, Ishikawa, Japan\\
$^{2}$Institute of Low Temperature Science, Hokkaido University, Kita-19, Nishi-8, Kita-ku, Sapporo, 060-0819, Hokkaido, Japan\\
$^{3}$Institute for Liberal Arts and Sciences, Kyoto University, 302 Konoe-kae, 69 Konoe-cho, Sakyo-ku, Kyoto, 606-8315, Kyoto, Japan\\
$^{4}$Institute for Chemical Reaction Design and Discovery (WPI-ICReDD), Hokkaido University, N21 W10, Kita-ku, Sapporo, 001-0021, Hokkaido, Japan\\
$^{*}$Corresponding author: katsuno@staff.kanazawa-u.ac.jp
\end{center}

\vspace{1em}

\section{Methods}
For the thickness of a solution sample in the liquid cell,
 the simplest thickness measurement method is to use electron energy loss spectroscopy (EELS), as reported \citep{holtz2013situ}.
Because EELS is available only for a thin liquid layer, we used another approach for the thickness measurement of a solution sample.
Specifically, we prepared a liquid cell with nanoparticles on the upper and lower thin membranes.
When the focus was adjusted onto a certain nanoparticle (labeled Q)
 and the stage was tilted,
 the position of a nanoparticle attached to the same membrane onto which nanoparticle Q was attached did not change.
By contrast, the position of particles attached to the opposite membrane changed.
By changing the tilting angle of the stage,
 we could derive the distance between the upper and lower membranes from the change in positions of the nanoparticles.

This idea can be used for the measurement of the thickness of thin films.
We prepared an elastic carbon grid with nanoparticles attached on both sides.
The elastic carbon grid consisted of a $20$--$25$\,nm carbon membrane and a $20$--$30$\,nm tri-acetylcellulose membrane.
The total thickness of the elastic carbon membrane is approximately $40$--$55$ nm; that is, the film thickness was less than the diameter of a nanoparticle of $100$\,nm we used.
Using the proposed method, we estimate the thickness of the film to be $58$ $\pm$ $33$\,nm on average for 10 samples.
This value is in good agreement with the product specifications.

\begin{figure}[bt]
\center
\includegraphics[width=\textwidth, bb=100 50 860 490]{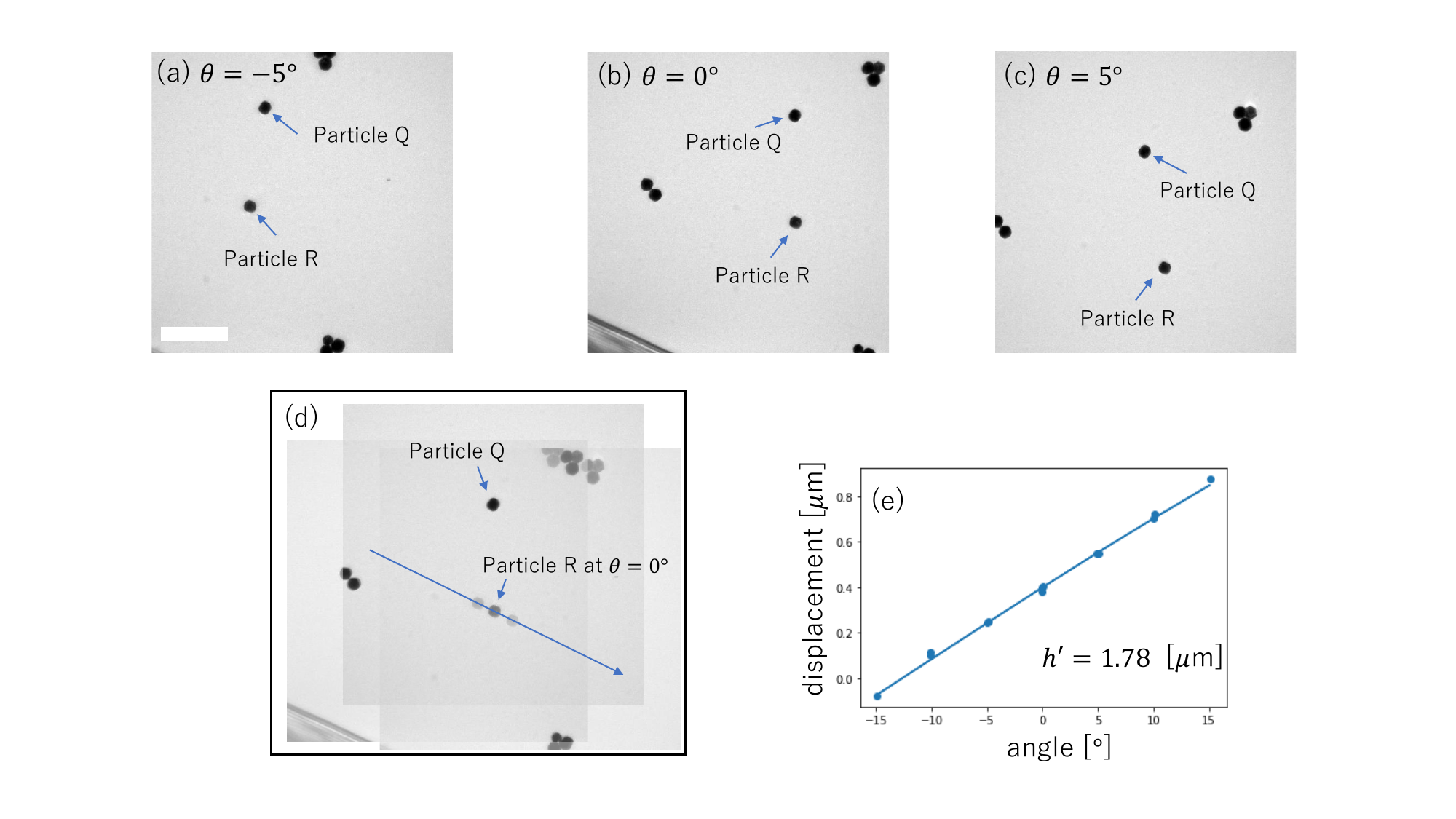}
\caption{
Examples of images for the measurement of the distance between upper and lower membranes in the absence of a solution.
Acquired images with the stage angle (a) $\theta=-5^\circ$, (b) $\theta=0^\circ$ and 
(c) $\theta=5^\circ$.
(d) The image merged from (a), (b), and (c).
When the stage is tilted, the position of out-of-focused particle R changes, 
whereas
 the position of focused particle Q does not.
(e) Angle dependence of the displacement $\ell(\theta)$ of particle R.
The origin of the displacement is the position of R at angle $\theta=0^{\circ}$.
The fitting function is $\ell(\theta) = h^{\prime}\sin(\theta)$.
The scale bar in (a) represents $500$\,nm.
}
\label{fig:fig/figs_p24}
\end{figure}

An example of images for the measurement of the distance between upper and lower membranes in the absence of a solution is shown in Figure~\ref{fig:fig/figs_p24}.
We used a liquid cell consisting of pairs of silicon chips with an amorphous silicon nitride membrane
 of $50$\,nm thickenss
 as an observation window and an LC-TEM holder (Poseidon, Protochips, Morrisville, NC, USA).
Au nanoparticles with a diameter of $100$\,nm were attached to the membranes inside the cell.
No liquid was injected, and the inside of the cell was exposed to the atmosphere
 through the injection ports of the TEM holder because the cell was not sealed.
Figures~\ref{fig:fig/figs_p24}a--c show the images
 acquired at tilting angles $-5^{\circ}$, $0^{\circ}$ and $5^{\circ}$, respectively.
When the three images are superimposed so that particle Q overlaps (Figure~\ref{fig:fig/figs_p24}d),
 the positions of some particles change with the angle, as shown by particle R in Figure~\ref{fig:fig/figs_p24}.
The displacement of particle R is plotted in Figure~\ref{fig:fig/figs_p24}e when the origin of the displacement is the position at the angle $\theta=0^{\circ}$.
The displacement $\ell$ depends on the tilting angle $\theta$.
Using the tilting angle $\theta$ and the distance between the top and bottom membrane $h^{\prime}$,
 we can write the displacement $\ell$ geometrically as
 $\ell(\theta) = h^{\prime}\sin(\theta)$.
Using the data of the displacement for some angles,
 we can obtain the distance between thin membranes from the least-squares method as $h^{\prime}$=1.78\,$\mu$m.
Because the nanoparticles with a diameter of $100$\,nm were attached inside the cell,
 the thickness of the liquid $h$ was $1.88$\,$\mu$m.

\begin{figure}[tb]
\center
\includegraphics[width=\textwidth, bb=0 50 960 490, clip]{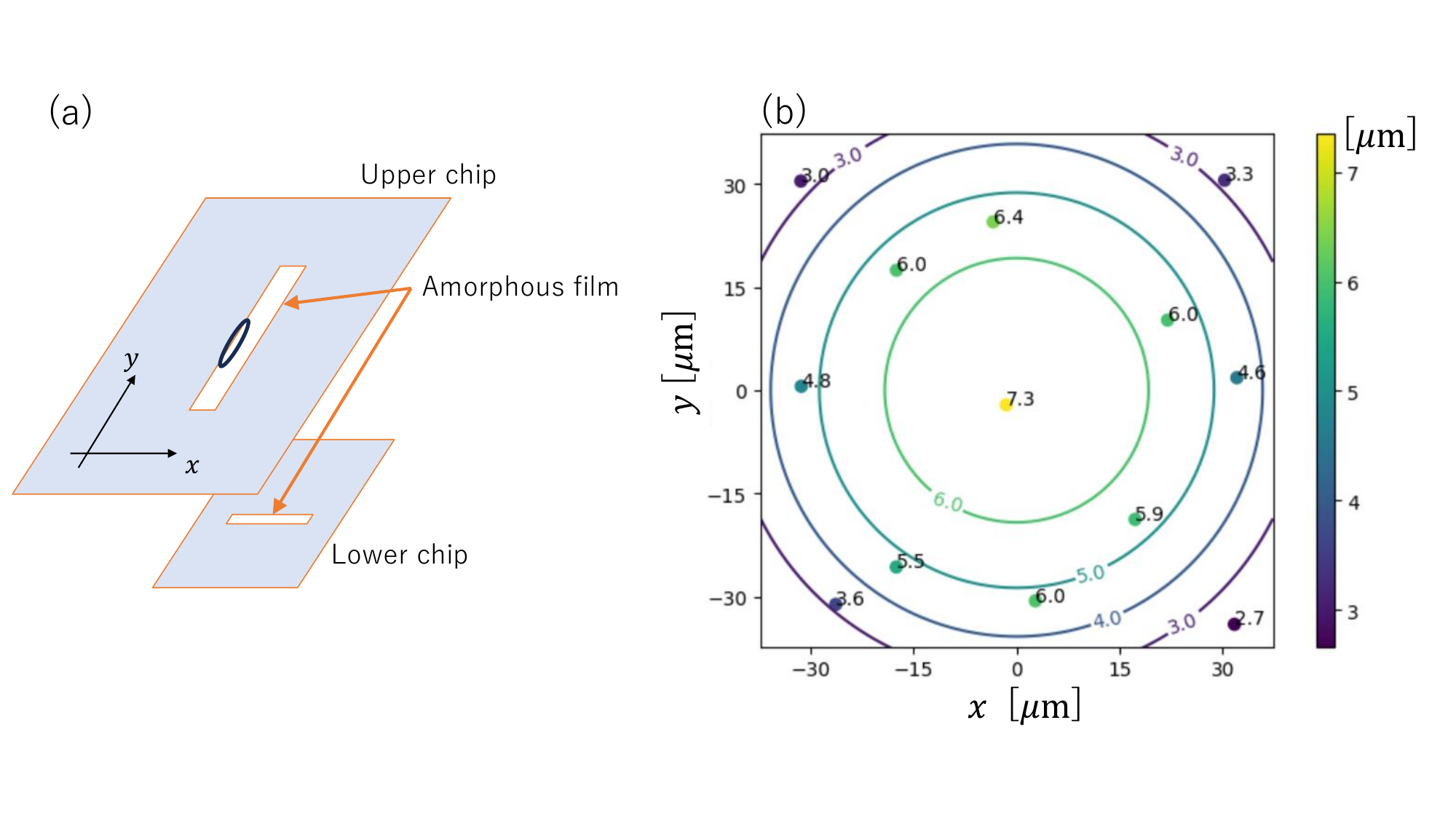}
\caption{
(a) Structure of a liquid cell. 
The thickness profile indicated by the ellipse is shown in Figure~\ref{fig:fig/figs_p33}.
There is an amorphous membrane in the center of each chip, and electron beams penetrate the membrane to make it visible.
(b) Liquid thickness profile with a $500$\,nm spacer and a $50$\,$\mu$m window size.
 Water as a solution was injected into the cell.
 The solid curves were obtained using equation~\ref{eqn:h}.
}
\label{fig:fig/figs_p25}
\end{figure}

\section{Results}
Four types of spacer sizes ($50$\,nm, $150$\,nm, $500$\,nm, and $5$\,$\mu$m) and two window sizes ($20$\,$\mu$m and  $50$\,$\mu$m) were prepared.
The window in the upper and lower chips is rectangular, and one side is approximately $500$\,$\mu$m.
The square window is formed by overlapping the crossed rectangular windows shown in Figure~\ref{fig:fig/figs_p25}a.
The cells were filled with water.

Using our proposed method,
 we measured the liquid thickness at thirteen points for each combination of spacers and window sizes.
Figure~\ref{fig:fig/figs_p25}b shows the result of the liquid thickness for a $500$\,nm spacer and a $50$\,$\mu$m window size in considering the nanoparticle size.
The actual window size is approximately $68$\,$\mu$m from the average length of the four window edges.
We find that the center of the window is bulging compared with the edges.
The thickness of the window edge reaches about $3$\,$\mu$m despite the use of a $500$\,nm spacer,
 and the thickness in the center reaches more than $7$\,$\mu$m.
The tendency of our results is consistent with that of the previously reported results \citep{holtz2013situ}.
Although the thickness lacks circular symmetry because of the structure of the cell,
 the thickness appears to be approximated by a quadratic function $h^{\prime}=-a r^{2} + b$ with distance from the center $r=\sqrt{x^{2}+y^{2}}$, where $x$ and $y$ represent the position.
The parameter $a$ is likely related to the elastic modulus of the membrane, and $b$ is the thickness at the center of the liquid.
After simple fitting, we find
 $a=2.3$\,mm$^{-1}$ and $b=7.5$\,$\mu$m
 from the experimentally obtained data.

We investigated the relation between the fitting parameters and the experimental setup.
Parameters $a$ and $b$ obtained for other combinations of spacer size $S$ and window size $W$ are summarized in Table~\ref{tab:tab1}.
The actual average window size of four edges $W_{\rm A}$ obtained from the measurement is listed for later discussion.
The actual window size was $1.5$ and $1.1$ times larger than the specifications of $20$\,$\mu$m window and $50$\,$\mu$m window, respectively.
The variation of the edge length is within approximately $20$\,\%, and we assume that the window shape is square.
The fitting parameters $a$ and $b$ appear to depend on the window size rather than on the spacer size.

\begin{table}[tb]
\center
\caption{
Fitting parameters $a$ and $b$ for the liquid thickness of a whole window.
$W$, $W_{\rm A}$, and $S$ represent the window size, actual window size, and spacer size, respectively.
The fitting function for the liquid thickness is $h^{\prime}=-ar^{2}+b$ where $r$ represents the distance from the center of the window.
}
\label{tab:tab1}
\begin{tabular}{ccccc}
\hline
$W$ [$\mu$m] & $W_{\rm A}$ [$\mu$m] & $S$ [nm] &  $a$ [mm$^{-1}]$ & $b$ [$\mu$m] \\
\hline
\hline
20  & 36 & 50   & 3.7 & 4.1\\
20  & 38 & 150  & 3.6 & 4.9\\
20  & 34 & 500  & 4.3 & 4.9\\
50  & 66 & 150  & 2.2 & 7.0\\
50  & 68 & 500  & 2.3 & 7.5\\
50  & 64 & 5000 & 2.1 & 10.1\\
\hline
\end{tabular}
\end{table}

\begin{table}[tb]
\center
\caption{
Fitting parameters $g_{\rm U}$, $g_{\rm L}$, $h_{\rm U}$, and $h_{\rm L}$ for the liquid thickness at the window edge.
The fitting function is $h^{\prime}=-g_{\rm L}x^{2}+h_{\rm L}$ or $h^{\prime}=-g_{\rm U}y^{2}+h_{\rm U}$.
}
\label{tab:tab2}
\begin{tabular}{ccccccc}
\hline
$W$ [$\mu$m] & $W_{\rm A}$ [$\mu$m] & $S$ [nm] &  $g_{\rm U}$ [mm$^{-1}$] & $g_{\rm L}$  [mm$^{-1}$] & $h_{\rm U}$ [$\mu$m] & $h_{\rm L}$ [$\mu$m] \\
\hline
\hline
20  & 36 & 50   & 7.5 & 2.8 & 1.5  & 0.6  \\
20  & 38 & 150  & 6.9 & 2.0 & 0.97 & 0.30 \\
20  & 34 & 500  & 6.8 & 2.9 & 1.4  & 0.40 \\
50  & 66 & 150  & 2.7 & 1.9 & 2.5  & 1.3  \\
50  & 68 & 500  & 3.4 & 1.5 & 3.1  & 1.6  \\
50  & 64 & 5000 & 2.8 & 1.1 & 2.0  & 0.81 \\
\hline
\end{tabular}
\end{table}

Because both the upper and lower membranes are bulging near the center of the membrane,
 it is difficult to specify the shape of each membrane from parameter $b$.
When the cell is examined along the edge of the observation window,
 one would expect to see one of the membranes nearly flattened.
That is, the thickness at the window edge along the $y$-direction is approximately described by $h^{\prime}=-g_{\rm L} y^{2} + h_{\rm L}$,
 where $g_{\rm L}$ and $h_{\rm L}$ represent the curvature of the lower membrane and the maximum thickness at the edge, respectively.
From the thickness at the window edge along the $x$-direction,
 the curvature of the upper membrane $g_{\rm U}$ and maximum thickness at the edge $h_{\rm U}$ are obtained in the same manner.
The fitting parameters are listed in Table~\ref{tab:tab2}.

We found that $1/g_{\rm U}$ and $1/g_{\rm L}$ are not sensitive to the spacer size and increase linearly with increasing window size.
Using our data, we can obtain $1/g_{\rm U, L} = \tilde{a}_{\rm U, L}W_{\rm A}$, where $\tilde{a}_{\rm U}$ and $\tilde{a}_{\rm L}$ are $0.20$ and $0.11$, respectively.

\begin{figure}[tb]
\center
\includegraphics[width=\textwidth, bb=100 180 760 360, clip]{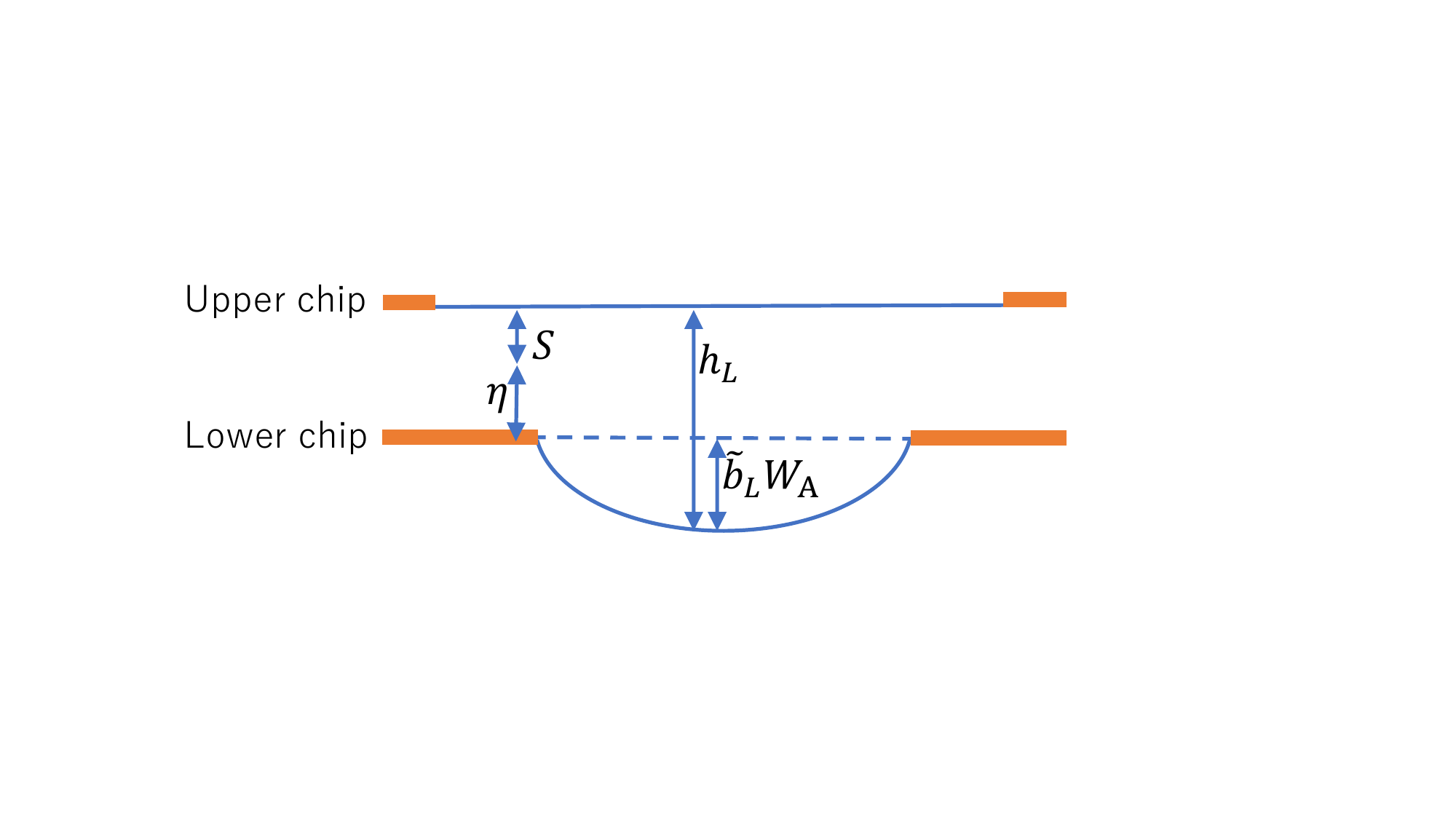}
\caption{
Schematics of the shape of a membrane on a lower chip, and the associated parameters.
The membrane of the upper chip is assumed to exhibit no deformation because the area is near the window edge (see Figure~\ref{fig:fig/figs_p25}a).
Thick orange solid lines represent chips, and thin solid lines represent membranes.
The broken line is a guide for the eyes.
}
\label{fig:fig/figs_p33}
\end{figure}

We also found that $h_{\rm U}$ and $h_{\rm L}$ are not sensitive to the spacer size and increase linearly with increasing window size.
If the window size is zero, then $h_{\rm U}$ and $h_{\rm L}$ would be expected to become equal to the spacer size $S$.
However, the results show that the experimentally obtained thickness has an unknown factor $\eta$, as shown in Figure~\ref{fig:fig/figs_p33}.
Our data show that 
 the maximum thickness at the window edge is expressed as $h_{\rm U,L} = \tilde{b}_{\rm U,L}W_{\rm A}+S+\eta$
 where $\tilde{b}_{\rm U}$, $\tilde{b}_{\rm L}$, and $\eta$ are obtained as $0.038$, $0.018$, and $2.5\,\mu$m, respectively.

If we assume that the membrane deforms with circular symmetry,
 we can obtain an empirical function of the liquid thickness in the liquid cell as
\begin{align}
h=-\frac{\sqrt{\tilde{a}_{\rm U}\tilde{a}_{\rm L}}}{W_{\rm A}} r^{2} + (\tilde{b}_{\rm U}+\tilde{b}_{\rm L})W_{\rm A} + \eta + S
\label{eqn:h}
\end{align}
where $\sqrt{\tilde{a}_{\rm U}\tilde{a}_{\rm L}}$ is related to the geometric average of the curvature of both membranes,
 $\tilde{b}_{\rm U}+\tilde{b}_{\rm L}$ is related to the total protrusion of the membrane at the center,
 and $\eta$ is an unknown value derived from our measurement.
Because of the finite size of the nanoparticles that we used in the measurement,
 it is necessary to subtract their contribution.
However, when the values $\sqrt{\tilde{a}_{\rm U}\tilde{a}_{\rm L}}$ and $\tilde{b}_{\rm U}+\tilde{b}_{\rm L}$ obtained from our measurement are used,
 the thickness estimation is not substantially affected because of the large effect of the unknown factor $\eta$.
In our experiments, those values are $\sqrt{\tilde{a}_{\rm U}\tilde{a}_{\rm L}}=0.15$, $\tilde{b}_{\rm U}+\tilde{b}_{\rm L} =0.056$, and $\eta=2.5\,\mu$m.
The empirical function equation~\ref{eqn:h} and those parameters reproduce the experimental data as shown in Figure~\ref{fig:fig/figs_p25}b.
From the definition of $\eta$,
 a small spacer ($\lesssim$ $2.5\,\mu$m) does not work.
Parameters $\tilde{a}_{\rm U}$ and $\tilde{a}_{\rm L}$ should be attributed to the properties of the material,
 and the magnitude of $\tilde{a}_{\rm U}$ is identical to that of $\tilde{a}_{\rm L}$ because the membrane is made of amorphous silicon nitride.
Although the reason is not clear, 
 these results are consistent with the observation that the upper membrane is more deformed from $b_{\rm U}>b_{\rm L}$.
In detail, the magnitude of parameters listed in Table~\ref{tab:tab2} for the upper membrane are also systematically two or three times larger than those for the lower membrane.
However, the finding that $g_{\rm U}$ and $g_{\rm L}$ depend on the window size $W_{\rm A}$ is unexpected.
The state of the amorphous membrane might vary depending on the size of the window.

Although there are some unknown facts about the shape of the membrane,
 we found that the thickness of the liquid is approximately twice as thick in spatial variation
 and that the thickness is effectively almost independent of the spacer.
The thickness of the liquid is approximately $3$--$7$\,$\mu$m in the case of $W=50$\,$\mu$m and $S=500$\,nm (Table~\ref{tab:tab1}).


\bibliographystyle{abbrvnat}